\def\year{2021}\relax
\documentclass[letterpaper]{article} % DO NOT CHANGE THIS
\usepackage{aaai21}  % DO NOT CHANGE THIS
\usepackage{times}  % DO NOT CHANGE THIS
\usepackage{helvet} % DO NOT CHANGE THIS
\usepackage{courier}  % DO NOT CHANGE THIS
\usepackage[hyphens]{url}  % DO NOT CHANGE THIS
\usepackage{graphicx} % DO NOT CHANGE THIS
\usepackage{natbib}  % DO NOT CHANGE THIS AND DO NOT ADD ANY OPTIONS TO IT
\usepackage{caption} % DO NOT CHANGE THIS AND DO NOT ADD ANY OPTIONS TO IT
\usepackage{soul}
\usepackage{url}
\usepackage{graphicx}
\usepackage{amsmath}
\usepackage{booktabs}
\usepackage{subfigure}
\usepackage{amsmath, amssymb}
\usepackage{makecell}
\usepackage{multirow}
\usepackage{footnote}
\usepackage{amssymb}
\usepackage{nameref}

\usepackage{times}
\usepackage{latexsym}
\usepackage[switch]{lineno}  %

\title{DUMA: Reading Comprehension with Transposition Thinking }
\author {
Pengfei Zhu,\textsuperscript{\rm 1,2,3}
Hai Zhao,\textsuperscript{\rm 1,2,3,\footnote{Corresponding author. This paper was partially supported by National Key Research and Development Program of China (No. 2017YFB0304100), Key Projects of National Natural Science Foundation of China (U1836222 and 61733011), Huawei-SJTU long term AI project, Cutting-edge Machine reading comprehension and language model.}}
Xiaoguang Li\textsuperscript{\rm 4} \\
}
\affiliations {
\textsuperscript{\rm 1}Department of Computer Science and Engineering, Shanghai Jiao Tong University\\
\textsuperscript{\rm 2}Key Laboratory of Shanghai Education Commission for Intelligent Interaction and Cognitive Engineering, Shanghai Jiao Tong University, Shanghai, China\\
\textsuperscript{\rm 3}MoE Key Lab of Artificial Intelligence, AI Institute, Shanghai Jiao Tong University, Shanghai, China\\
\textsuperscript{\rm 4}Huawei Noah's Ark Lab\\
}

\begin{document}

\maketitle

\begin{abstract}
Multi-choice Machine Reading Comprehension (MRC) requires model to decide the correct answer from a set of answer options when given a passage and a question. Thus in addition to a powerful Pre-trained Language Model (PrLM) as encoder, multi-choice MRC especially relies on a matching network design which is supposed to effectively capture the relationships among the triplet of passage, question and answers. While the newer and more powerful PrLMs have shown their mightiness even without the support from a matching network, we propose a new \textbf{DU}al \textbf{M}ulti-head Co-\textbf{A}ttention (DUMA) model, which is inspired by human's transposition thinking process solving the multi-choice MRC problem: respectively considering each other's focus from the standpoint of passage and question. The proposed DUMA has been shown effective and is capable of generally promoting PrLMs. Our proposed method is evaluated on two benchmark multi-choice MRC tasks, DREAM and RACE, showing that in terms of powerful PrLMs, DUMA can still boost the model to reach new state-of-the-art performance.

\end{abstract}

\section{Introduction}
\begin{figure}
	\centering
	\includegraphics[width=1\linewidth]{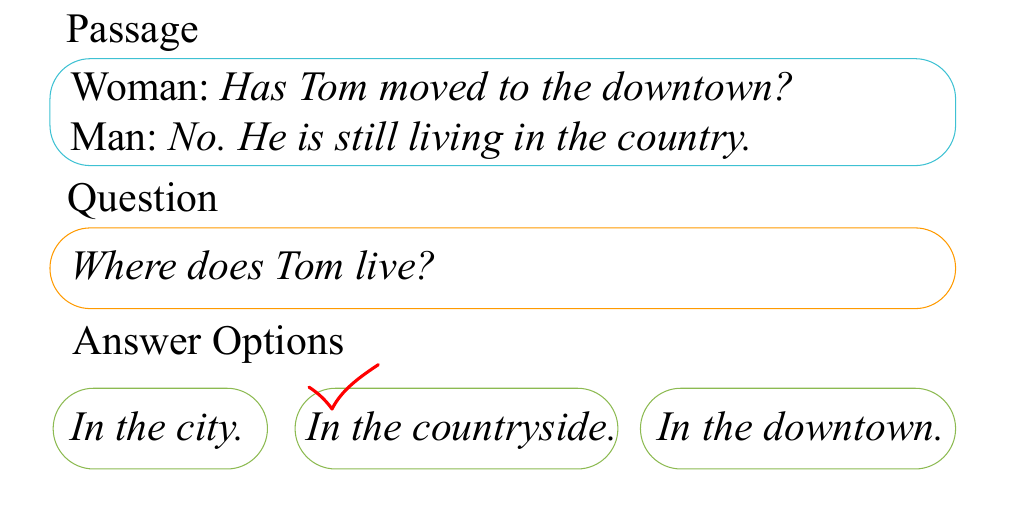}
	\caption{\label{tab:example_dream} An example of DREAM dataset.}
\end{figure}

%mrc-attention-transformer-预训练模型-mrc任务特点-结构性差异
Machine Reading Comprehension has been a heated topic and challenging problem, and various datasets and models have been proposed in recent years \cite{newsqa,ecawe,msmarco,squad,teach_machine_read,dream,race,dua,lingke,zhang2020sg,bhargav2020translucent,hu2019read+}. For the tasks of MRC, given passage and question, the task can be categorized as \textit{generative} and \textit{selective} according to its answer style \cite{mrc_survey}. \textit{Generative} tasks require the model to generate answers according to the passage and question, not limited to spans of the passage, while \textit{selective} tasks give model several candidate answers to select the best one. Multi-choice MRC is a typical task in \textit{selective} type, xwhich is the focus of this paper. Figure \ref{tab:example_dream} shows one example of DREAM dataset \cite{dream}, whose task is to select the best answer among three candidates given particular passage and question. %The task of representative dataset DREAM is to select the best answer among three candidates given particular passage and question.
 
The kernel method for a model to solve MRC problem is a two-level hierarchical process, 1) representation encoding which is done by an encoder such as PrLM; and 2) capturing the relationship among the triplet of passage, question and answer which has to be carefully handled by various matching networks such as OCN \cite{ocn} and DCMN \cite{dcmn}. With the development of PrLMs, matching network design tends to become more complicated for more effective improvements. 

\makesavenoteenv{tabular}
\makesavenoteenv{table}
\begin{table}[h]\small
\renewcommand\arraystretch{1.3}
	\centering
	{
		\begin{tabular}{@{}p{1.9cm}|@{}p{0.45cm}|l|l|p{1.3cm}}
			\hline		
			 		& & +OCN & +WAE & +DCMN \\
			\hline
			\hline
			BERT$_{\rm large}$  & \,71.0 & 71.7(+0.7)$^*$ & 73.1(+2.1)& 75.8(+4.8)$^*$ \\
			XLNet$_{\rm large}$& \,80.1 & 80.9(+0.8) & 81.8(+1.7)& 82.8(+2.7)$^*$ \\
			ALBERT$_{\rm xxlarge}$  & \,86.6  & 87.2(+0.6) & 87.3(+0.7) & 85.7(-0.9) \\
			ELECTRA$_{\rm large}$  &  \,86.1 & 86.3(+0.2) & 86.9(+0.8)&  84.9(-1.2)\\
			\hline
		\end{tabular}
		
	}
	\caption{Improvements of several prior models for representative PrLMs (sorted by releasing time) on RACE dataset.}
	\label{lm_and_net_compare}
\end{table}

Table \ref{lm_and_net_compare} shows that as the newer variant of the PrLM such as ALBERT \cite{albert} has shown its powerfulness even without the support from a proper matching network, in the meantime, the previous models\footnote{We re-implement OCN and WAE, and obtain codes of DCMN through personal communication with its authors. Besides, results denoted with $^*$ are from original papers.} \cite{ocn, bert_wae, dcmn} either brings very limited improvements or even cause drop on the PrLM's \cite{bert, xlnet, albert, clark2020electra} performance, which motivates us to develop an more effective mechanism to support the powerful enough PrLMs. Instead of designing more complicated matching network patterns, we choose a going-back-to-the-basic way to have obtained inspiration from human experience on solving MRC problems, which intuitively is to first \textbf{1)} quickly read through the overall content of passage, question and answer options to build up a global impression, followed by a \textbf{transposition thinking} process: \textbf{2)} based on dedicated information from question and answer options, re-considerate details of the passage and collect supporting evidences for question and answer options, \textbf{3)} based on dedicated information from passage, re-considerate the question and answer options to decide the correct option and exclude wrong options. When humans are re-reading the passage, they tend to extract key information according to their impression of question and answer options, and it is the same when re-reading question and answer options. It can be regarded as a bi-directional process in terms of transposition thinking pattern, and we adopt an attention inside network design to simulate this procedure, whose details are shown in the following Section \textit{\nameref{sec_model}}.

Since attention mechanism was proposed \cite{nmt_attention} originally for Neural Machine Translation, it has been widely used in MRC tasks to model the relationships between passage and question, and effectively enhances nearly all kinds of tasks \cite{bidaf,dua,dcmn}. Attention mechanism computes relationships of each word representation in one sequence to a target word representation in another sequence and aggregates them to form a final representation, which is commonly named as passage-to-question attention or question-to-passage attention.

Transformer \cite{transformer} uses self-attention mechanism to represent dependencies and relationships of different positions in one single sequence, which is an effective method to obtain representations of sentences for global encoding. Since \cite{gpt,bert} use it to improve the structure of PrLMs \cite{elmo}, many kinds of PrLMs has been proposed to constantly refresh records of all kinds of tasks \cite{roberta,albert}. For PrLMs, the more layer and bigger hidden size they use, the better performance they achieve. Benefited from large-scale unlabeled training data and multiple stacked layers, PrLMs are able to encode sentences into very deep and precise representations. Moreover, \cite{albert} reveals the importance of generalization for models, that is parameter sharing among layers can efficiently improve the performance. However, training a LM has been a time and labor consuming work, which usually needs amounts of engineering works to explore parameter settings. The bigger the model is, the more resource it consumes and the harder it can be implemented. Moreover, despite the great success they achieve in different tasks, we find that for MRC tasks, using self-attention of the Transformer to model sequences is far from enough. No matter how deep the structure is, it suffers from the nature of self-attention, which is only drawing a global relationship, while for MRC tasks the passage and the question are remarkable different in contents and literal structures and the relationship between them necessarily needs to be carefully considered. However, previous models \cite{nmt_attention,bidaf,dcmn} only obtain limited improvement when applied on the top of PrLMs even though they use very complicated structure.

Rather than seeking a complicated matching network pattern, we are inspired by the human thinking experience solving MRC problems and put forward a new network design named as \textbf{DU}al \textbf{M}ulti-head Co-\textbf{A}ttention (DUMA) to sufficiently capture relationships among passage, question and answer options for multi-choice MRC, as a result it may effectively improve the performance when cooperating with newer and more powerful PrLMs. Our model is based on the Multi-head Attention module, which is the kernel module of Transformer. Similar to BiDAF \cite{bidaf} and DCMN \cite{dcmn}, we use the bi-directional way to obtain sufficient modeling of relationships. The contributions can be summarized as:

1) For multi-choice MRC tasks, we investigate effects of previous models over Pre-trained Language Models.

2) We propose a new \textbf{DU}al \textbf{M}ulti-head Co-\textbf{A}ttention (DUMA) model which well simulates the procedure human solving MRC tasks, and show its effectiveness and superiority to previous models through extensive experiments.

3) We have reached new state-of-the-art on two benchmark multi-choice MRC tasks, DREAM and RACE.

\section{Related Works}\label{related}
\cite{nmt_attention} first propose attention mechanism for Neural Machine Translation. The jointly learning of alignment and translation effectively improves the performance. Since then, attention model has been introduced to all kinds of Natural Language Processing tasks and various of architectures has been proposed \cite{tu2020select,chen2019convolutional,gao2019generating,yan2019deep}. \cite{bidaf} uses a multi-stage architecture to hierarchically model representation of the passage, and uses a bi-directional attention flow. These works are before PrLMs was proposed, and are able to model the representations well on the top of traditional encoder such as Long Short-Term Memory (LSTM) \cite{lstm}. In fact, the experimental results show that they can still improve the representations of PrLMs, but the improvements are suboptimal. 

Based on PrLMs, \cite{ocn} propose a method to model relationship and interaction among answer options to the benefit of distinguishing them. \cite{bert_wae} ensemble a model which learns to select the wrong answer. \cite{dcmn} propose a sentence selection method to select more important sentences from passage to improve the matching representations, and considers interactions among answers for multi-choice MRC tasks. Even though the matching network design becomes more complicated, it cannot fully exploit powerful PrLMs and even cause drop on performance when applied on newer PrLMs\footnote{As shown in Table \ref{lm_and_net_compare}.}.

In a word, when applied on the top of PrLMs, previous models are not effective enough to improve the performance by a large margin. Thus inspired by the experience human solving MRC problems we design a new model which can effectively utilize well-modeled representations of PrLMs for even better performance.

\section{Task Definition}
Multi-choice MRC tasks have to handle a triplet of passage $P$, question $Q$ and answer $A$. When given the passage and question, the model is required to make a correct answer. The passage consists of multiple sentences, and its content can be dialogue, story, news and so on, depending on the domain of the dataset. The questions and corresponding answers are single sentences, which are usually much shorter than the passage. The target of multi-choice MRC is to select the correct answer from the candidate answer set $ A = \{A_{1}, ... , A_{t}\} $ for a given passage and question pair $<P,Q>$, where $t$ is the number of candidate answers. Formally, the model needs to learn a probability distribution function $F(A_1,A_2,...,A_t|P,Q)$.

\begin{figure}[!htb]
	\centering
	\includegraphics[width=0.45 \textwidth]{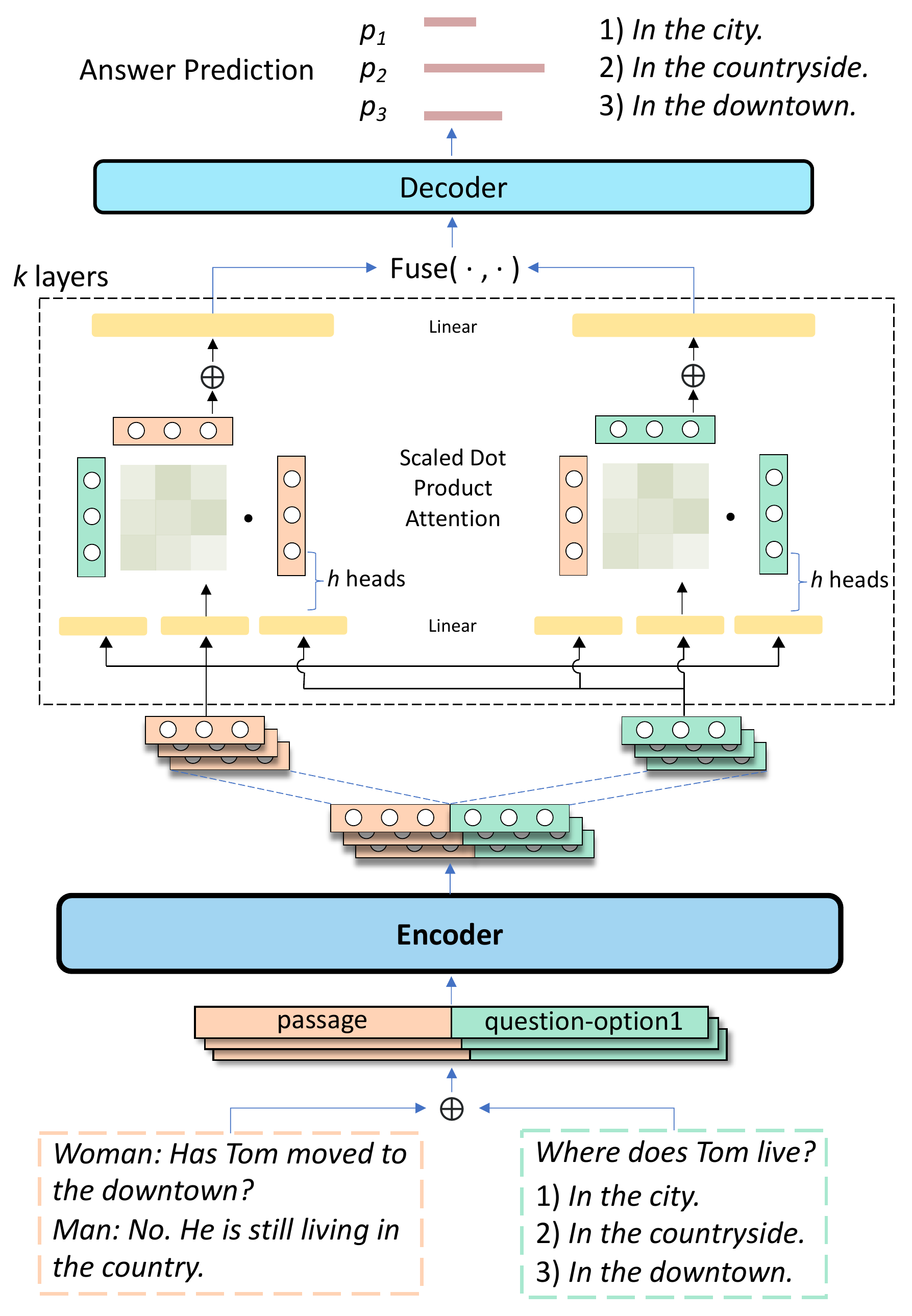}
	\caption{The overall architecture. Our proposed DUMA is between the Encoder and Decoder.}
	\label{fig:overview}
\end{figure}

\section{Model}
\label{sec_model}

Figure \ref{fig:overview} illustrates the overall architecture of our model. An encoder takes text input to form a global sequence representation, which is similar to human reading through the whole content for the first time to obtain an overall impression, and a decoder is to perform the answer prediction which is similar to human aggregating all the information to select the correct answer option. Our proposed Dual Multi-head Co-Attention (DUMA) layer is between the encoder and the decoder, which simulates human transposition thinking process to capture relationships of key information from passage, question and answer options.

\subsection{Encoder}
To encode input tokens into representations, we take PrLMs as the encoder. To get global contextualized representation, for each different candidate answer, we concatenate it with its corresponding passage and question to form one sequence and then feed it into the encoder. Let $ P = [p_{1}, p_{2}, ... , p_{m}] $, $ Q = [q_{1}, q_{2}, ... , q_{n}] $, $ A = [a_{1}, a_{2}, ... , a_{k}] $ respectively denote the sequences of passage, question and a candidate answer, where $p_{i}$, $q_{i}$, $a_{i}$ are tokens. The adopted encoder with encoding function $Enc(\cdot)$ takes the concatenation of $P$, $Q$ and $A$ as input, namely $E=Enc(P\oplus Q\oplus A)$. The encoding output $E$ has a form $ [e_1, e_2, ..., e_{m+n+k}]$, where $e_i$ is a vector of fixed dimension $d_{model}$ that represents the respective token.

\subsection{Dual Multi-head Co-Attention}

%\begin{figure}\centering
%	\subfigure[]{
%		\begin{minipage}[t]{0.4\linewidth}
%			\includegraphics[width=1\textwidth]{mha.pdf}\label{tab:mha}
%		\end{minipage}
%	}
%	\subfigure[]{
%		\begin{minipage}[t]{0.45\linewidth}
%			\includegraphics[width=1\textwidth]{mhca.pdf}\label{tab:mhca}
%		\end{minipage}
%	}
%	\caption{(a) Original Multi-head Attention. (b) Our Dual Multi-head Co-attention. The two Multi-head Attention module share parameters.} 
%\end{figure} 

We use our proposed Dual Multi-head Co-Attention module to calculate attention representations of passage and question-answer. Figure \ref{fig:overview} shows the details of our proposed DUMA, which may be stacked as $k$ layers. The following formula takes $k=1$ for simplicity. Our model is based on the Multi-head Attention module \cite{transformer}. The proposed DUMA reuses the architecture of Multi-head Attention, while for the inputs, $K$ and $V$ are the same but $Q$ is another sequence representation (Note this $Q$ here denotes \textit{Query} from the original paper, different from previous $Q$ in this paper. And $K$, $V$ are \textit{Key}, \textit{Value} respectively). We first separate the output representation from Encoder to obtain $E^P = [e_{1}^p, e_{2}^p, ..., e_{l_{p}}^p]$ and $E^{QA} = [e_{1}^{qa}, e_{2}^{qa}, ..., e_{l_{qa}}^{qa}]$, where $e_{i}^p$, $e_{j}^{qa}$ denote the $i$-th and $j$-th token representation of passage and question-answer respectively and $l_p$, $l_{qa}$ are the length. Then we calculate the attention representations in a bi-directional way, that is, take 1) $E^P$ as \textit{Query}, $E^{QA}$ as \textit{Key} and \textit{Value}, and 2) $E^{QA}$ as \textit{Query}, $E^{P}$ as \textit{Key} and \textit{Value}.
\begin{equation}\small
	\begin{split}
	{\rm Attention}(E^P,E^{QA},E^{QA})&={\rm softmax}(\frac{E^P{(E^{QA})}^T}{\sqrt{d_k}})E^{QA}\\
	{\rm head_i} = {\rm Attention}(E^P&W_i^Q, E^{QA}W_i^K, E^{QA}W_i^V) \\
	{\rm MHA}(E^P,E^{QA},E^{QA})=&{\rm Concat}({\rm head_1}, ..., {\rm head_h})W^O \\
	{\rm MHA_1=MHA}&(E^P, E^{QA}, E^{QA}) \\
	{\rm MHA_2=MHA}&(E^{QA}, E^P, E^P) \nonumber
	\end{split}
\end{equation}
\begin{equation}\small
	{\rm DUMA}(E^P,E^{QA}) = {\rm Fuse}(\rm {MHA_1, MHA_2})\label{eq_fuse}\\ 
\end{equation}

\noindent where $W_i^Q \in \mathbb{R}^{d_{model} \times d_q}$, $W_i^K \in \mathbb{R}^{d_{model} \times d_k}$, $W_i^V \in \mathbb{R}^{d_{model} \times d_v}$, $W_i^O \in \mathbb{R}^{hd_{v} \times d_{model}}$ are parameter matrices, $d_q$, $d_k$, $d_v$ denote the dimension of \textit{Query} vectors, \textit{Key} vectors and \textit{Value} vectors, $h$ denotes the number of heads, $MHA(\cdot)$ denotes Multi-head Attention and $DUMA(\cdot)$ denotes our Dual Multi-head Co-Attention. The $Fuse(\cdot,\cdot)$ function first uses mean pooling to pool the sequence outputs of $MHA(\cdot)$, and then aggregates the two pooled outputs through a fusing method. In Subsection \textit{\nameref{fuse_investigate}}, we investigate three fusing methods, namely element-wise multiplication, element-wise summation and concatenation. 

As shown in the Figure \ref{fig:overview}, the left part of DUMA calculates question-answer-aware passage representation, which simulates human re-reading details in the passage with impression of question and answer, and the right part calculates passage-aware question-answer representation, which simulates re-considering the question-answer with deeper understanding of the passage. The $Fuse(\cdot,\cdot)$ function means fusing all the key information before deciding which is the best answer option.

\subsection {Decoder}
Our model decoder takes the outputs of DUMA and computes the probability distribution over answer options. Let $A_i$ denote the $i$-th answer option, $O_i \in \mathbb{R}^l $ denote the output of $i$-th $<P,Q,A_i>$ triplet, and $A_r$ denote the correct answer option, the loss function is computed as:
\begin{equation}
\begin{split}
	&O_i = {\rm DUMA}(E^P,E^{QA_i}) \\
	&L(A_r|P,Q) = -{\rm log}\frac{{\rm exp}(W^TO_r)}{\sum^s_{i=1}{\rm exp}(W^TO_i)} \nonumber
	\end{split}
\end{equation}
where $W \in \mathbb{R}^l $ is a learnable parameter and $s$ denotes the number of candidate answer options.

\section{Experiments}
\begin{table}[t]\small
\renewcommand\arraystretch{1.1}
	\centering
	{
		\begin{tabular}{l|l|l}
			\hline		
			  & DREAM & RACE  \\
			\hline
			\hline
			\# of source documents & 6,444 & 27,933    \\
			\# of total questions & 10,197 & 97,687 \\
			Train/Dev/Test split & 3:1:1 &18:1:1 \\
			Extractive (\%)   & 16.3 & 13.0   \\
			Abstractive (\%)  & 83.7  & 87.0 \\
			Average answer length  & 5.3 & 5.3  \\
			\# of answers per question & 3 & 4 \\
			Avg./Max. \# of turns per dialogue & 4.7 / 48 & -  \\
			Avg. passage length & 85.9 & 321.9  \\
			Vocabulary size  & 13,037 & 136,629 \\
			\hline
		\end{tabular}
		
	}
	\caption{\label{statistics} Statistical data of DREAM and RACE dataset. \# denotes the number. ``Extractive" means the answers are spans of the passage, and ``Abstractive" means the answers are not spans.}
\end{table}

Our proposed method is evaluated on two benchmark multi-choice MRC tasks, DREAM and RACE. Table \ref{statistics} shows their data statistics, which indicates RACE is a large-scale dataset covering a broad range of domains, and DREAM is a small dataset presenting passage in a form of dialogue.

\paragraph{DREAM} DREAM \cite{dream} is a dialogue-based dataset for multiple-choice reading comprehension, which is collected from English exams. Each dialogue as the given passage has multiple corresponding questions and each question has three candidate answers. The most important feature of the dataset is that most of the questions are non-extractive and need reasoning from more than one sentence, so the dataset is small but still challenging. 
\paragraph{RACE} RACE \cite{race} is a large dataset collected from middle and high school English exams. Most of the questions also need reasoning, and domains of passages are diversified, ranging from news, story to ads.

\subsection{Evaluation}
For multi-choice MRC tasks, the evaluation criteria is accuracy, $acc = N^+ /  N$, where $N^+$ denotes the number of examples the model selects the correct answer, and $N$ denotes the number of the whole evaluation examples.

\begin{table}[t]\small
\renewcommand\arraystretch{1.3}
	\centering
	{
		\begin{tabular}{p{4.4cm}|@{}p{0.45cm}|@{}p{0.45cm}|@{}p{1.3cm}}
			\hline		
			 model &\;dev & \;test &\;source  \\
			\hline
			\hline
			BERT$_{\rm large}$\cite{bert}  & \,66.0 & \,66.8 &\multirow{6}*{\,leaderboard}   \\
			BERT$_{\rm large}$+WAE& \multirow{2}*{\,-} & \multirow{2}*{\,69.0}   \\
			\cite{bert_wae} &&\\
			XLNet$_{\rm large}$\cite{xlnet} & \,-  & \,72.0 \\
			RoBERTa$_{\rm large}$+MMM& \multirow{2}*{\,88.0} & \multirow{2}*{\,88.9}\\ \cite{mmm} &&  \\
			\hline
			ALBERT$_{\rm xxlarge}$\cite{albert} & \,89.2  & \,88.5 \\
			\hline
			ALBERT$_{\rm xxlarge}$+DUMA & \textbf{\,89.9} & \textbf{\,90.5}& \multirow{2}*{\,our\;\;model} \\
			\;\;\;+multi-task learning\cite{dumamulti}&\,- & \textbf{\,91.8} \\
			\hline
		\end{tabular}
	}
	\caption{\label{tab:dream_result} Results on DREAM dataset. Results with multi-task learning are reported by \cite{dumamulti}.}
\end{table}

\begin{table}[t]\small
\renewcommand\arraystretch{1}
	\centering
	{
		\begin{tabular}{p{3.3cm}|p{1.9cm}|@{}p{1.2cm}}
			\hline		
			 model  & test (M/H) & \;source  \\
			\hline
			\hline
			HAF \cite{haf}  &46.0(45.0/46.4) & \multirow{10}*{\;publication}   \\
			MRU \cite{mru}    & 50.4(57.7/47.4)   \\
			HCM \cite{hcm}   & 50.4(55.8/48.2) \\
			MMN \cite{mmn}  & 54.7(61.1/52.2)  \\
			GPT \cite{gpt} & 59.0(62.9/57.4)  \\
			RSM \cite{rsm} & 63.8(69.2/61.5)  \\
			OCN \cite{ocn} & 71.7(76.7/69.6) \\
			XLNet \cite{xlnet} & 81.8(85.5/80.2) \\
			XLNet$_{\rm xxlarge}$ + DCMN+ & \multirow{2}*{82.8(86.5/81.3)} \\
			\cite{dcmn} & \\
			\hline
%			XLNet + DCMN+ & 82.8(86.5/81.3) & \multirow{10}*{leaderboard}\\
%			RoBERTa & 83.2(86.5/81.8) &~\\
%			DCMN+ (ensemble) & 84.1(88.5/82.3) &~\\
			RoBERTa + MMM & \multirow{2}*{85.0(89.1/83.3)} & \multirow{13}*{\;leaderboard}\\
			\cite{mmm}&&\\
			ALBERT (single) & \multirow{2}*{86.5(89.0/85.5)} &~\\
			\cite{albert} &&\\
			T5$^*$\cite{t5} & 87.1(-/-)&~ \\
			UnifiedQA &\multirow{2}*{89.4(-/-)} &~ \\
			\cite{unifiedqa}&&\\
			ALBERT(ensemble) & \multirow{2}*{89.4(91.2/88.6)} & ~  \\
			\cite{albert} &&\\
			Megatron-BERT (single) & \multirow{2}*{\textbf{89.5(91.8/88.6)}} &~ \\
			\cite{Megatron}&&\\
			Megatron-BERT (ensemble)\cite{Megatron} & \multirow{2}*{\textbf{90.9(93.1/90.0)}} &\multirow{2}*{~} \\
			\cline{1-3}
			ALBERT$_{\rm xxlarge}$ & \multirow{2}*{86.6(89.0/85.5)}\\
			 \cite{albert}  &  \\
			\hline
			ALBERT$_{\rm xxlarge}$+DUMA & \textbf{88.0(90.9/86.7)} & \multirow{3}*{\;our model} \\
			ALBERT$_{\rm xxlarge}$+DUMA&\multirow{2}*{\textbf{89.8(92.6/88.7)}}\\
			(ensemble) &  &  \\
			\hline
		\end{tabular}
		
	}
	\caption{\label{tab:race_result} Results on RACE dataset.}
\end{table}

\begin{table}[h]\small
\renewcommand\arraystretch{1.3}
	\centering
	{
		\begin{tabular}{l|l|l}
			\hline		
			 model & dev & test (M/H)  \\
			\hline
			\hline
			ALBERT$_{\rm xxlarge}$  & 87.4  & 86.6(89.0/85.5) \\
			\hline
			ALBERT$_{\rm xxlarge}$&\multirow{2}*{\textbf{88.1}(+0.7)}&\multirow{2}*{\textbf{88.0(90.9/86.7)}(+1.4)}\\
			+DUMA &  &  \\
			\hline
		\end{tabular}
		
	}
	\caption{\label{tab:race_result_1} Comparison with ALBERT baseline on RACE dataset.}
\end{table}

\begin{table}[t]\small
\renewcommand\arraystretch{1.1}
	\centering
	{
		\begin{tabular}{l|l|l}
			\hline		
			 model & dev & test  \\
			\hline
			\hline
			ALBERT$_{\rm base}$ & 64.51 & 64.43	 \\
			\hline
			\;\;+Vanilla SA&66.27&66.34\\
			\;\;+DUMA & 67.06 & \textbf{67.56}\\
			\;\;+TB-DUMA & \textbf{67.79} & 67.17 \\
			\hline
		\end{tabular}
		
	}
	\caption{\label{trm_and_mha} Comparison among vanilla Multi-head Self-attention, DUMA and TB-DUMA on DREAM dataset.}
\end{table}

\subsection{Experimental Settings}
Our model takes ALBERT$_{{\rm xxlarge}}$ as encoder, and use $k=2$ layers of DUMA. We make the left and right part of DUMA and all the layers share parameters. Using the PrLM, our model training is done through a fine-tuning way for both tasks.

Our codes are written based on Transformers\footnote{https://github.com/huggingface/transformers}, and results of ALBERT \cite{albert}, ELECTRA \cite{clark2020electra} and BERT \cite{bert} models as baselines are our re-running unless otherwise specified.

For DREAM dataset, the learning rate is 1e-5, batch size is 8 and the warmup steps are 100. We train the model for 2 epochs in 4 hours. For RACE dataset, the learning rate is 1e-5, the batch size is 8 and the warmup steps are 1000. We train the model for 3 epochs in 2 days. For each dataset, we use FP16 training from Apex\footnote{https://github.com/NVIDIA/apex} for accelerating the training process. We train the models on eight nVidia P40 GPUs. In the following Section \textit{\nameref{ablation_sec}}, for other re-running or re-implementation including PrLM baselines and PrLM plus other models for comparison, we use the same learning rate, warmup steps and batch size as mentioned above. 

We choose the result on dev set that has stopped increasing for three checkpoints (382 steps for DREAM and 3000 steps for RACE). To obtain stable results, we run experiments 5 times with different random seeds and select the median as the ultimate performance.

\subsection{Results}
Tables \ref{tab:dream_result}, \ref{tab:race_result} and \ref{tab:race_result_1} show the experimental results. Megatron-BERT \cite{Megatron} is a variant of BERT \cite{bert} which has 8.3 billion parameters and is nearly 40 times bigger than the largest size of ALBERT, so usually it is very hard applied in practice with present common computation power and its results are not strictly comparable to our ALBERT+DUMA. Except for this, our model both achieves state-of-the-art performance on RACE leaderboard\footnote{http://www.qizhexie.com/data/RACE\_leaderboard} and DREAM leaderboard\footnote{https://dataset.org/dream/}, and it can be further improved with multi-task learning method MMM \cite{dumamulti, mmm}.

\section{Analysis Studies}\label{ablation_sec}

We perform ablation experiments on the DREAM dataset to investigate key features of our proposed DUMA, such as attention modeling ability, structural simplicity, bi-directional setting and low coupling.

\subsection{Comparison with Vanilla Self-attention and Transformer Block}\label{trm_block}

We investigate whether the improvements are simply caused by the increase of parameters. Thus we conduct the experiments of ALBERT plus vanilla Multi-head Self-attention \cite{transformer}, whose inputs $Q$, $K$, $V$ are all concatenation of passage, question and answer. Results shown in Table \ref{trm_and_mha} indicate the effectiveness of our bi-directional co-attention model design.

Moreover, we observe that the original Transformer Block (TB) \cite{transformer} consists not only \textit{Multi-head Attention} module but also \textit{Layer Normalization (LN)} and \textit{Feed-Forward Network (FFN)}.  In consideration of the extensive application and great success of TB for global encoding \cite{bert,roberta,albert}, we investigate whether the Transformer Block better model the co-attention relationships than Multi-head Attention using TB-based DUMA (TB-DUMA). Experimental results shown in Table \ref{trm_and_mha} indicate TB-DUMA has no obvious difference with our DUMA in modeling relationships. However, our proposed DUMA holds more brief structure and equally effective performance.

\subsection{Comparison with Related Models}
\begin{table}[t]\small
\renewcommand\arraystretch{1.3}
	\centering
	{
		\begin{tabular}{@{}p{2.65cm}|l|p{1.85cm}}
			\hline		
			 model & ALBERT$_{\rm base/xxlarge}$  & ELECTRA$_{\rm large}$   \\
			\hline
			\hline
			baseline & 64.4/88.5 & 88.2	 \\
			\hline
			+Soft\,Attention\shortcite{nmt_attention}& 65.4(+1.0)/88.9(+0.4) & 88.8(+0.6) \\
			+BiDAF\shortcite{bidaf} & 65.6(+1.2)/89.3(+0.8) & 89.1(+0.9)\\
			+OCN\shortcite{ocn} &65.8(+1.4)/89.2(+0.7)  & 89.0(+0.8) \\
			+WAE\shortcite{bert_wae} & 66.5(+2.1)/89.9(+1.4) & 89.5(+1.3) \\
			
			+DCMN\footnote{The results of ALBERT+DCMN are our re-running of the official codes which we obtained through personal communication with its authors.}\shortcite{dcmn} & 63.3(-1.1)/87.8(-0.7)  & 87.7(-0.5)\\
			\hline
			+DUMA & \textbf{67.6(+3.2)/90.5(+2.0)} & \textbf{89.8(+1.6)}  \\
			\hline
		\end{tabular}
	}
	\caption{\label{attention_comparison} Comparison among different models on DREAM dataset.}
\end{table}

We compare our attention model with several representative works, which have been discussed in Section \textit{\nameref{related}}. Soft Attention \cite{nmt_attention} and BiDAF \cite{bidaf} are originally based on traditional encoder such as LSTM \cite{lstm}, and DCMN \cite{dcmn}, OCN \cite{ocn}, WAE\cite{bert_wae} are based on BERT \cite{bert}. For fair comparison with Soft Attention, we simply use it to replace the attention score computing in our model. 

Table \ref{attention_comparison} compares the effectiveness of various model designs, and our proposed DUMA outperforms all other models. The performance of Soft Attention is much lower than our DUMA, which indicates the DUMA's similarity in structure with ALBERT (both use Multi-head Attention) makes it better to utilize information from encoded representation. Even though BiDAF has been a successful attention model since a long time ago, it is suboptimal for PrLMs. WAE uses an ensemble model design with nearly twice sized parameters as our model. DCMN adopts a much more complicated model structure design for better matching, but the result with ALBERT and ELECTRA is not satisfactory, which indicates it may be specially optimized for specific PrLM, while our DUMA achieves the absolutely highest accuracy with a intuitive structure design. In fact, our DUMA has nice generalization ability because it also works well with many kinds of PrLMs.

\subsection{Investigation of Fusing Method} \label{fuse_investigate}

\begin{table}[t]\small
\renewcommand\arraystretch{1.3}
	\centering
	{
		\begin{tabular}{l|l|l}
			\hline		
			 model & dev & test  \\
			\hline
			\hline
			ALBERT$_{\rm base}$ & 64.51 & 64.43	 \\
			\hline
			element-wise multiplication & 65.29 &64.58\\
			element-wise summation & 66.32 & 65.51 \\
			concatenation & \textbf{67.06} & \textbf{67.56}\\
			\hline
		\end{tabular}
		
	}
	\caption{\label{fusing_methods} Comparison among different implementation of the fusing method on DREAM dataset. The last three rows are our DUMA applying three kinds of implementations.}
\end{table}

We investigate different implementations of fusing function from equation (\ref{eq_fuse}), namely element-wise multiplication, element-wise summation and concatenation. The results are shown in Table \ref{fusing_methods}. We see that concatenation is optimal because it retains the matching information and lets network learn to fuse them dynamically.

\subsection{Number of Parameters}
\begin{table}[t]\small
\renewcommand\arraystretch{1.3}
	\centering
	{
		\begin{tabular}{p{3.15cm}|p{3.2cm}}
			\hline		
			 model & para. num.  \\
			\hline
			\hline
			ALBERT$_{\rm base}$ & 11.7M	 \\
			\hline
			\;+Soft Attention\shortcite{nmt_attention} & 13.5M\;(+1.8M)\;(+15.4\%)   \\
			\;+BiDAF\shortcite{bidaf} &  12.0M\;(+0.3M)\;(+2.6\%) \\
			\;+OCN\shortcite{ocn} & 14.8M\;(+3.1M)\;(+26.5\%)  \\
			\;+WAE\shortcite{bert_wae} & 23.4M\;(+11.7M)\;(+100\%)  \\
			\;+DCMN\shortcite{dcmn} & 19.4M\;(+7.7M)\;(+65.8\%)  \\
			\hline
			\;+DUMA & 13.5M\;(+1.8M)\;(+15.4\%) \\
			\hline
		\end{tabular}
		
	}
	\caption{\label{parameter_comparison} Comparison of number of parameters among different models. The models are same as listed in Table \ref{attention_comparison}.}
\end{table}
We compare number of parameters among different models in Table \ref{parameter_comparison}. BiDAF requires the least model enlargement, however it is far less effective than our model. Besides, our model enlargement is far less than DCMN. In a word, our DUMA can obtain the best performance while requiring a little model enlargement.

%\footnotetext[1]{The results of ALBERT+DCMN are our re-running, instead of re-implementations, of official code, which we obtained through personal communication with its authors.}

\begin{figure}[h]\small\centering
	\subfigure[]{
		\begin{minipage}[]{0.45\linewidth}
			\includegraphics[width=1\textwidth]{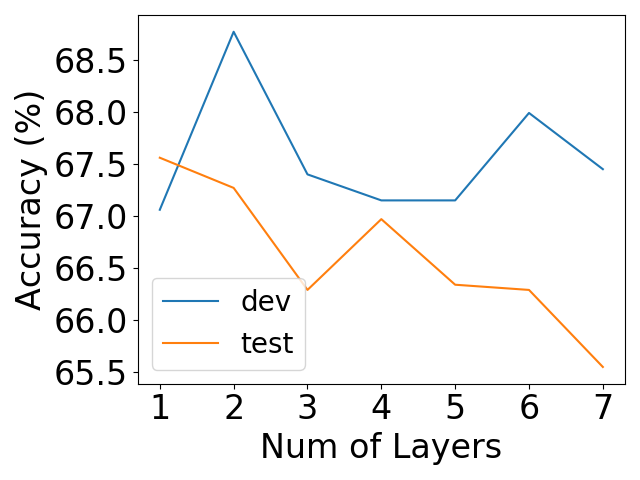}\label{multi_layer}
		\end{minipage}
	}
	\subfigure[]{
		\begin{minipage}[]{0.45\linewidth}
			\includegraphics[width=1\textwidth]{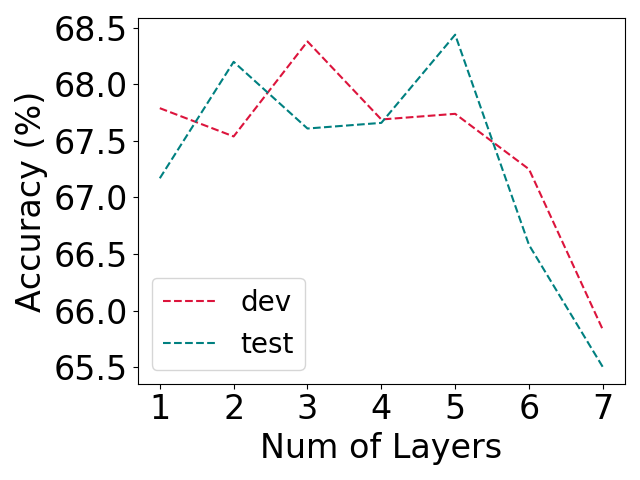}\label{multi_layer_1}
		\end{minipage}
	}
	\caption{(a) Different numbers of DUMA layers on DREAM dataset. (b) Different numbers of TB-DUMA layers on DREAM dataset.} 
\end{figure} 
\subsection{Number of DUMA Layers}

%\begin{figure}
%	\centering
%	\includegraphics[width=0.4\textwidth]{multi_layer.png}
%	\caption{\label{multi_layer} Different numbers of DUMA layers on DREAM dataset.}
%\end{figure}
%\begin{figure}
%	\centering
%	\includegraphics[width=0.4\textwidth]{multi_layer_1.png}
%	\caption{\label{multi_layer_1} Different numbers of SAN-DUMA layers on DREAM dataset.}
%\end{figure}

We stack 2 layers of our DUMA, that is to make passage and question-answer interact more than once to obtain deeper representations. Besides, we make different layers share parameters, which is the same as ALBERT. 

Figure \ref{multi_layer} shows the results. We can see that as the number of layers increases the performance fluctuates, and too many layers even lead to slight drop. It is much like when human solving MRC tasks, excessive thinking and hesitation may make them misunderstand the meaning of some information. For the network with current number of parameters, it shows that interacting twice is enough to capture the key information, and stacking too many layers may disorder the well-modeled representations and make the model harder trained. 

Note that PrLMs \cite{bert,roberta,albert} stacks Transformer Blocks (described in Subsection \textit{\nameref{trm_block}}) instead of Multi-head Attention modules, which raises a doubt that it is the lack of LN and FFN that makes the DUMA improper for stacking deeper network. Thus we further conduct an experiment with TB-DUMA (the same as described in Subsection \textit{\nameref{trm_block}}). Experimental results in Table \ref{multi_layer_1} show the same performance trend as the original DUMA, which again verifies the effectiveness of our DUMA design. 

\subsection{Effect of Bi-direction}
\begin{table}[t]\small
\renewcommand\arraystretch{1.1}
	\centering
	{
		\begin{tabular}{@{}p{1.7cm}|p{1.6cm}|p{1.6cm}|p{1.6cm}}
			\hline		
			 \;model & dev & test & avg \\
			\hline
			\hline
			\;ALBERT$_{\rm base}$ & 64.51 & 64.43 &64.47 \\
			\hline
			\;P-to-Q & 64.61\;(+0.10) & 64.72\;(+0.29) & 64.67\;(+0.20) \\
			\;Q-to-P & 66.76\;(+2.25) & 66.29\;(+1.86) & 66.53\;(+2.06)\\
			\;Both(DUMA) & 67.06\;(\textbf{+2.55}) & 67.56\;(\textbf{+3.13}) & 67.31\;(\textbf{+2.84})\\
			\hline
		\end{tabular}
		
	}
	\caption{\label{single_direction} Bi-directional vs. uni-directional attentions on DREAM dataset.}
\end{table}

As figured out by \cite{dcmn}, bi-directional matching is a very important feature for sufficiently modeling the relationship between passage and question. To investigate the effect, we perform experiments on two settings, namely P-to-Q only and Q-to-P only. In other words, we respectively remove the right part and left part of our DUMA. Table \ref{single_direction} shows the results. We see that for bi-directional model, the overall improvement is 2.84\%, while for uni-directional model the improvement is only 2.06\% at most. The setting of bi-direction effectively improves the performance, which reveals its efficiency for modeling the relationship and agrees to the conclusion of \cite{dcmn}. Also it is the same as our intuitive understanding that all the passage, question and answer options should be deliberated.

\subsection{Cooperation with PrLMs} \label{with_bert}
\begin{table}[t]\small
\renewcommand\arraystretch{1.2}
	\centering
	{
		\begin{tabular}{p{1.6cm}|p{1.6cm}|p{1.6cm}|p{1.6cm}}
			\hline		
			 model & dev & test &avg \\
			\hline
			\hline
			ALBERT$_{\rm base}$ & 64.51 & 64.43 &64.47 \\
			\;\;+DUMA &  67.06\;(+2.55) & 67.56\;(+3.13) & 67.31\;(+2.84)\\
			\hline
			BERT$_{\rm base}$ & 61.18  &  61.54 & 61.36 \\
			\;\;+DUMA & 64.82\;(+3.64) & 64.03\;(+2.49) & 64.43\;(+3.07) \\
			\hline
		\end{tabular}
		
	}
	\caption{\label{bert_result} Results using BERT as encoder on DREAM dataset. Results of BERT$_{\rm base}$ are our re-running.}
\end{table}

Though the proposed DUMA is supposed to enhance state-of-the-art PrLM like ALBERT and ELECTRA, we claim that it is generally effective for less advanced models. Thus we simply replace the adopted ALBERT by its early variant BERT to examine the effectiveness of DUMA. Table \ref{bert_result} shows the results. We see that our model can be easily transferred to other PrLMs, thus it can be seemed as an effective module for modeling relationships among passage, question and answer for Multi-choice MRC.

\subsection{Effect of Self-attention}
\begin{table}[t]\small
\renewcommand\arraystretch{1.1}
	\centering
	{
		\begin{tabular}{l|l|l}
			\hline		
			 model & dev & test  \\
			\hline
			\hline
			ALBERT$_{\rm base}$ (SA) & 64.51 & 64.43 \\
			ALBERT$_{\rm base}$ (SA) + DUMA (CA) & \textbf{67.06} & \textbf{67.56} \\
			ALBERT$_{\rm base}$ (SA+CA) & 41.18  &  40.08 \\
			\hline
		\end{tabular}
		
	}
	\caption{\label{only_co_attention} Results with and without self-attention on DREAM dataset. ``SA" means self-attention and ``CA" means co-attention. ``SA+CA" means straightforwardly using CA to replace SA in ALBERT.}
\end{table}
Our overall architecture can be split into two steps from the view of attention, of which the first is self-attention (ALBERT) and the second is co-attention (DUMA). To examine whether the structure can be further simplified, that is only using co-attention, we straightforwardly change all of the Multi-head Self-attention of ALBERT model to our Dual Multi-head Co-attention, while still using its pre-trained parameters. The results are shown in Table \ref{only_co_attention}, showing that putting co-attention directly into ALBERT model may lead to much poorer performance compared to the original ALBERT and our ALBERT+DUMA integration way. To conclude, a better way for modeling is our PrLM plus DUMA model, which is to firstly build a global relationship using self-attention of the well trained encoder and then further enhance the relationship between passage and question-answer and distill more matching information using co-attention.
\section{Comparison of Predictions}

Table \ref{prediction_comparison} shows a hard example which needs to capture important relationships and matching information. Benefited from well-modeled relationship representations, DUMA can better distill important matching information between passage and question-answer.

\begin{table}[h]\small
\renewcommand\arraystretch{1.3}
	\centering
	\begin{tabular}{p{1.2cm}|p{1.1cm}|p{1.0cm}|p{1.0cm}|p{1.0cm}}
		\hline
		\multirow{4}*{Passage} & \multicolumn{4}{l}{Woman: \textit{So, you have three days off, what}}\\ 
		& \multicolumn{4}{l}{\textit{are you going to do?}} \\
		& \multicolumn{4}{l}{Man: \textit{Well, I probably will rent some movi-}} \\
		& \multicolumn{4}{l}{\textit{es with my friend bob.}} \\
		\hline
		Question & \multicolumn{4}{l}{\textit{What will the man probably do?}} \\
		\hline
		\multirow{3}{1cm}{Answer options}& \multicolumn{4}{l}{1) \textit{Ask for a three-day leave.}} \\ 
		& \multicolumn{4}{l}{2) \textit{Go out with his friend.}} \\ 
		& \multicolumn{4}{l}{3) \textit{Watch films at home.}\;\;\;$\surd$} \\
		\hline
		ALBERT&+BiDAF&+Sf Att&+DCMN&+DUMA \\
		\hline
		Prediction&\multicolumn{1}{c|}{1)}&\multicolumn{1}{c|}{1)}&\multicolumn{1}{c|}{2)}&\multicolumn{1}{c}{3)\;\;$\surd$}\\
		\hline
	\end{tabular}
	\caption{\label{prediction_comparison} Predictions of different models which are same as in Table \ref{attention_comparison}. ``Sf Att'' means Soft Attention.}
	\end{table}
\section{Conclusion}

In this paper, we simulates human transposition thinking experience when solving MRC problems and propose a novel \textbf{DU}al \textbf{M}ulti-head Co-\textbf{A}ttention (DUMA) to model the relationships among passage, question and answer for multi-choice MRC tasks, which is able to cooperate with popular large-scale Pre-trained Language Models and brings effective performance improvements. Besides, we investigate previous attention mechanisms or matching networks applied on the top of PrLMs, and our model is shown as optimal through extensive experiments, which achieves the best performance with an intuitive motivated structure design. Our proposed DUMA enhancement has been verified effective on two benchmark multi-choice MRC tasks, DREAM and RACE, which achieves new state-of-the-art over strong PrLM baselines.

\bibliography{aaai2021_arxiv.bib}

\begin{thebibliography}{41}
\providecommand{\natexlab}[1]{#1}
\providecommand{\url}[1]{\texttt{#1}}
\providecommand{\urlprefix}{URL }
\expandafter\ifx\csname urlstyle\endcsname\relax
  \providecommand{\doi}[1]{doi:\discretionary{}{}{}#1}\else
  \providecommand{\doi}{doi:\discretionary{}{}{}\begingroup
  \urlstyle{rm}\Url}\fi

\bibitem[{Bahdanau, Cho, and Bengio(2015)}]{nmt_attention}
Bahdanau, D.; Cho, K.; and Bengio, Y. 2015.
\newblock Neural Machine Translation by Jointly Learning to Align and
  Translate.
\newblock In \emph{3rd International Conference on Learning Representations
  ({ICLR} 2015)}.

\bibitem[{Baradaran, Ghiasi, and Amirkhani(2020)}]{mrc_survey}
Baradaran, R.; Ghiasi, R.; and Amirkhani, H. 2020.
\newblock A Survey on Machine Reading Comprehension Systems.
\newblock \emph{arXiv preprint arXiv:2001.01582} .

\bibitem[{Bhargav et~al.(2020)Bhargav, Glass, Garg, Shevade, Dana, Khandelwal,
  Subramaniam, and Gliozzo}]{bhargav2020translucent}
Bhargav, G.~S.; Glass, M.; Garg, D.; Shevade, S.; Dana, S.; Khandelwal, D.;
  Subramaniam, L.~V.; and Gliozzo, A. 2020.
\newblock Translucent Answer Predictions in Multi-Hop Reading Comprehension.
\newblock In \emph{Proceedings of the AAAI Conference on Artificial
  Intelligence}, volume~34, 7700--7707.

\bibitem[{Chen et~al.(2019)Chen, Cui, Ma, Wang, and Hu}]{chen2019convolutional}
Chen, Z.; Cui, Y.; Ma, W.; Wang, S.; and Hu, G. 2019.
\newblock Convolutional spatial attention model for reading comprehension with
  multiple-choice questions.
\newblock In \emph{Proceedings of the AAAI Conference on Artificial
  Intelligence}, volume~33, 6276--6283.

\bibitem[{Clark et~al.(2020)Clark, Luong, Le, and Manning}]{clark2020electra}
Clark, K.; Luong, M.-T.; Le, Q.~V.; and Manning, C.~D. 2020.
\newblock Electra: Pre-training text encoders as discriminators rather than
  generators.
\newblock \emph{arXiv preprint arXiv:2003.10555} .

\bibitem[{Devlin et~al.(2018)Devlin, Chang, Lee, and Toutanova}]{bert}
Devlin, J.; Chang, M.-W.; Lee, K.; and Toutanova, K. 2018.
\newblock {BERT}: Pre-training of Deep Bidirectional Transformers for Language
  Understanding.
\newblock \emph{arXiv preprint arXiv:1810.04805} .

\bibitem[{Gao et~al.(2019)Gao, Bing, Li, King, and Lyu}]{gao2019generating}
Gao, Y.; Bing, L.; Li, P.; King, I.; and Lyu, M.~R. 2019.
\newblock Generating distractors for reading comprehension questions from real
  examinations.
\newblock In \emph{Proceedings of the AAAI Conference on Artificial
  Intelligence}, volume~33, 6423--6430.

\bibitem[{Hermann et~al.(2015)Hermann, Kocisk{\'y}, Grefenstette, Espeholt,
  Kay, Suleyman, and Blunsom}]{teach_machine_read}
Hermann, K.~M.; Kocisk{\'y}, T.; Grefenstette, E.; Espeholt, L.; Kay, W.;
  Suleyman, M.; and Blunsom, P. 2015.
\newblock Teaching Machines to Read and Comprehend.
\newblock In \emph{Advances in Neural Information Processing Systems 28: Annual
  Conference on Neural Information Processing Systems 2015 ({NIPS} 2015)}.

\bibitem[{Hochreiter and Schmidhuber(1997)}]{lstm}
Hochreiter, S.; and Schmidhuber, J. 1997.
\newblock Long Short-Term Memory.
\newblock \emph{Neural Computation} 9(8): 1735--1780.
\newblock \doi{10.1162/neco.1997.9.8.1735}.
\newblock \urlprefix\url{https://doi.org/10.1162/neco.1997.9.8.1735}.

\bibitem[{Hu et~al.(2019)Hu, Wei, Peng, Huang, Yang, and Li}]{hu2019read+}
Hu, M.; Wei, F.; Peng, Y.; Huang, Z.; Yang, N.; and Li, D. 2019.
\newblock Read+ verify: Machine reading comprehension with unanswerable
  questions.
\newblock In \emph{Proceedings of the AAAI Conference on Artificial
  Intelligence}, volume~33, 6529--6537.

\bibitem[{Jin et~al.(2020)Jin, Gao, Kao, Chung, and Hakkani-tur}]{mmm}
Jin, D.; Gao, S.; Kao, J.-Y.; Chung, T.; and Hakkani-tur, D. 2020.
\newblock {MMM}: Multi-stage Multi-task Learning for Multi-choice Reading
  Comprehension.
\newblock In \emph{The Thirty-Fourth {AAAI} Conference on Artificial
  Intelligence ({AAAI} 2020)}.

\bibitem[{Khashabi et~al.(2020)Khashabi, Khot, Sabharwal, Tafjord, Clark, and
  Hajishirzi}]{unifiedqa}
Khashabi, D.; Khot, T.; Sabharwal, A.; Tafjord, O.; Clark, P.; and Hajishirzi,
  H. 2020.
\newblock Unified{QA}: Crossing format boundaries with a single {QA} system.
\newblock \emph{arXiv preprint arXiv:2005.00700} .

\bibitem[{Kim and Fung(2020)}]{bert_wae}
Kim, H.; and Fung, P. 2020.
\newblock Learning to Classify the Wrong Answers for Multiple Choice Question
  Answering (Student Abstract) .

\bibitem[{Lai et~al.(2017)Lai, Xie, Liu, Yang, and Hovy}]{race}
Lai, G.; Xie, Q.; Liu, H.; Yang, Y.; and Hovy, E. 2017.
\newblock {RACE}: Large-scale ReAding Comprehension Dataset From Examinations.
\newblock In \emph{Proceedings of the 2017 Conference on Empirical Methods in
  Natural Language Processing ({EMNLP} 2017)}.

\bibitem[{Lan et~al.(2020)Lan, Chen, Goodman, Gimpel, Sharma, and
  Soricut}]{albert}
Lan, Z.; Chen, M.; Goodman, S.; Gimpel, K.; Sharma, P.; and Soricut, R. 2020.
\newblock {ALBERT}: A Lite {BERT} for Self-supervised Learning of Language
  Representations.
\newblock In \emph{8th International Conference on Learning Representations
  ({ICLR} 2020)}.

\bibitem[{Liu et~al.(2019)Liu, Ott, Goyal, Du, Joshi, Chen, Levy, Lewis,
  Zettlemoyer, and Stoyanov}]{roberta}
Liu, Y.; Ott, M.; Goyal, N.; Du, J.; Joshi, M.; Chen, D.; Levy, O.; Lewis, M.;
  Zettlemoyer, L.; and Stoyanov, V. 2019.
\newblock Ro{BERT}a: A Robustly Optimized {BERT} Pretraining Approach.
\newblock \emph{arXiv preprint arXiv:1907.11692} .

\bibitem[{Nguyen et~al.(2016)Nguyen, Rosenberg, Song, Gao, Tiwary, Majumder,
  and Deng}]{msmarco}
Nguyen, T.; Rosenberg, M.; Song, X.; Gao, J.; Tiwary, S.; Majumder, R.; and
  Deng, L. 2016.
\newblock MS MARCO: A Human Generated MAchine Reading COmprehension Dataset.
\newblock In \emph{Proceedings of the Workshop on Cognitive Computation:
  Integrating neural and symbolic approaches 2016 co-located with the 30th
  Annual Conference on Neural Information Processing Systems}.

\bibitem[{Peters et~al.(2018)Peters, Neumann, Iyyer, Gardner, Clark, Lee, and
  Zettlemoyer}]{elmo}
Peters, M.; Neumann, M.; Iyyer, M.; Gardner, M.; Clark, C.; Lee, K.; and
  Zettlemoyer, L. 2018.
\newblock Deep contextualized word representations.
\newblock In \emph{Proceedings of the 2018 Conference of the North American
  Chapter of the Association for Computational Linguistics: Human Language
  Technologies ({NAACL-HLT} 2018)}.

\bibitem[{Radford et~al.(2018)Radford, Narasimhan, Salimans, and
  Sutskever}]{gpt}
Radford, A.; Narasimhan, K.; Salimans, T.; and Sutskever, I. 2018.
\newblock Improving language understanding with unsupervised learning.
\newblock \emph{Technical report, OpenAI.} .

\bibitem[{Raffel et~al.(2019)Raffel, Shazeer, Roberts, Lee, Narang, Matena,
  Zhou, Li, and Liu}]{t5}
Raffel, C.; Shazeer, N.; Roberts, A.; Lee, K.; Narang, S.; Matena, M.; Zhou,
  Y.; Li, W.; and Liu, P.~J. 2019.
\newblock Exploring the limits of transfer learning with a unified text-to-text
  transformer.
\newblock \emph{arXiv preprint arXiv:1910.10683} .

\bibitem[{Rajpurkar et~al.(2016)Rajpurkar, Zhang, Lopyrev, and Liang}]{squad}
Rajpurkar, P.; Zhang, J.; Lopyrev, K.; and Liang, P. 2016.
\newblock SQuAD: 100,000+ Questions for Machine Comprehension of Text.
\newblock In \emph{Proceedings of the 2016 Conference on Empirical Methods in
  Natural Language Processing ({EMNLP} 2016)}.

\bibitem[{Ran et~al.(2019)Ran, Li, Hu, and Zhou}]{ocn}
Ran, Q.; Li, P.; Hu, W.; and Zhou, J. 2019.
\newblock Option Comparison Network for Multiple-choice Reading Comprehension.
\newblock \emph{arXiv preprint arXiv:1903.03033} .

\bibitem[{Seo et~al.(2017)Seo, Kembhavi, Farhadi, and Hajishirzi}]{bidaf}
Seo, M.; Kembhavi, A.; Farhadi, A.; and Hajishirzi, H. 2017.
\newblock Bidirectional Attention Flow for Machine Comprehension.
\newblock In \emph{5th International Conference on Learning Representations
  ({ICLR} 2017)}.

\bibitem[{Shoeybi et~al.(2019)Shoeybi, Patwary, Puri, LeGresley, Casper, and
  Catanzaro}]{Megatron}
Shoeybi, M.; Patwary, M.; Puri, R.; LeGresley, P.; Casper, J.; and Catanzaro,
  B. 2019.
\newblock Megatron-{LM}: Training Multi-Billion Parameter Language Models Using
  Model Parallelism.
\newblock \emph{arXiv preprint arXiv:1909.08053} .

\bibitem[{Sun et~al.(2019{\natexlab{a}})Sun, Yu, Chen, Yu, Choi, and
  Cardie}]{dream}
Sun, K.; Yu, D.; Chen, J.; Yu, D.; Choi, Y.; and Cardie, C. 2019{\natexlab{a}}.
\newblock {DREAM:} {A} Challenge Dataset and Models for Dialogue-Based Reading
  Comprehension.
\newblock \emph{{TACL}} 7: 217--231.
\newblock
  \urlprefix\url{https://transacl.org/ojs/index.php/tacl/article/view/1534}.

\bibitem[{Sun et~al.(2019{\natexlab{b}})Sun, Yu, Yu, and and}]{rsm}
Sun, K.; Yu, D.; Yu, D.; and and, C.~C. 2019{\natexlab{b}}.
\newblock Improving Machine Reading Comprehension with General Reading
  Strategies.
\newblock In \emph{Proceedings of the 2019 Conference of the North American
  Chapter of the Association for Computational Linguistics: Human Language
  Technologies ({NAACL-HLT} 2019)}.

\bibitem[{Tang, Cai, and Zhuo(2019)}]{mmn}
Tang, M.; Cai, J.; and Zhuo, H.~H. 2019.
\newblock Multi-Matching Network for Multiple Choice Reading Comprehension.
\newblock In \emph{The Thirty-Third {AAAI} Conference on Artificial
  Intelligence ({AAAI} 2019)}.

\bibitem[{Tay, Tuan, and Hui(2018)}]{mru}
Tay, Y.; Tuan, L.~A.; and Hui, S.~C. 2018.
\newblock Multi-range Reasoning for Machine Comprehension.
\newblock \emph{arXiv preprint arXiv:1803.09074} .

\bibitem[{Trischler et~al.(2017)Trischler, Wang, Yuan, Harris, Sordoni,
  Bachman, and Suleman}]{newsqa}
Trischler, A.; Wang, T.; Yuan, X.; Harris, J.; Sordoni, A.; Bachman, P.; and
  Suleman, K. 2017.
\newblock NewsQA: A machine comprehension dataset.
\newblock In \emph{Proceedings of the 2nd Workshop on Representation Learning
  for NLP}.

\bibitem[{Tu et~al.(2020)Tu, Huang, Wang, Huang, He, and Zhou}]{tu2020select}
Tu, M.; Huang, K.; Wang, G.; Huang, J.; He, X.; and Zhou, B. 2020.
\newblock Select, Answer and Explain: Interpretable Multi-Hop Reading
  Comprehension over Multiple Documents.
\newblock In \emph{AAAI}, 9073--9080.

\bibitem[{Vaswani et~al.(2017)Vaswani, Shazeer, Parmar, Uszkoreit, Jones,
  Gomez, Kaiser, and Polosukhin}]{transformer}
Vaswani, A.; Shazeer, N.; Parmar, N.; Uszkoreit, J.; Jones, L.; Gomez, A.~N.;
  Kaiser, L.; and Polosukhin, I. 2017.
\newblock Attention Is All You Need.
\newblock In \emph{Advances in Neural Information Processing Systems 30: Annual
  Conference on Neural Information Processing Systems 2017 (NIPS 2017)}.

\bibitem[{Wan(2020)}]{dumamulti}
Wan, H. 2020.
\newblock Multi-task Learning with Multi-head Attention for Multi-choice
  Reading Comprehension.
\newblock \emph{arXiv preprint arXiv:2003.04992} .

\bibitem[{Wang et~al.(2018)Wang, Yu, Chang, and Jiang}]{hcm}
Wang, S.; Yu, M.; Chang, S.; and Jiang, J. 2018.
\newblock A Co-Matching Model for Multi-choice Reading Comprehension.
\newblock In \emph{Proceedings of the 56th Annual Meeting of the Association
  for Computational Linguistics ({ACL} 2018)}.

\bibitem[{Yan et~al.(2019)Yan, Xia, Wu, Bi, Zhao, Zhang, Si, Wang, Wang, and
  Chen}]{yan2019deep}
Yan, M.; Xia, J.; Wu, C.; Bi, B.; Zhao, Z.; Zhang, J.; Si, L.; Wang, R.; Wang,
  W.; and Chen, H. 2019.
\newblock A deep cascade model for multi-document reading comprehension.
\newblock In \emph{Proceedings of the AAAI Conference on Artificial
  Intelligence}, volume~33, 7354--7361.

\bibitem[{Yang et~al.(2019)Yang, Dai, Yang, Carbonell, Salakhutdinov, and
  Le}]{xlnet}
Yang, Z.; Dai, Z.; Yang, Y.; Carbonell, J.; Salakhutdinov, R.; and Le, Q.~V.
  2019.
\newblock {XLN}et: Generalized Autoregressive Pretraining for Language
  Understanding.
\newblock \emph{arXiv preprint arXiv:1906.08237} .

\bibitem[{Zhang et~al.(2020{\natexlab{a}})Zhang, Zhao, Wu, Zhang, Zhou, and
  Zhou}]{dcmn}
Zhang, S.; Zhao, H.; Wu, Y.; Zhang, Z.; Zhou, X.; and Zhou, X.
  2020{\natexlab{a}}.
\newblock {DCMN}+: Dual Co-Matching Network for Multi-choice Reading
  Comprehension.
\newblock In \emph{The Thirty-Fourth {AAAI} Conference on Artificial
  Intelligence ({AAAI} 2020)}.

\bibitem[{Zhang et~al.(2018{\natexlab{a}})Zhang, Huang, Zhu, and Zhao}]{ecawe}
Zhang, Z.; Huang, Y.; Zhu, P.; and Zhao, H. 2018{\natexlab{a}}.
\newblock Effective Character-augmented Word Embedding for Machine Reading
  Comprehension.
\newblock In \emph{Natural Language Processing and Chinese Computing - 7th
  {CCF} International Conference ({NLPCC} 2018)}.

\bibitem[{Zhang et~al.(2018{\natexlab{b}})Zhang, Li, Zhu, Zhao, and Liu}]{dua}
Zhang, Z.; Li, J.; Zhu, P.; Zhao, H.; and Liu, G. 2018{\natexlab{b}}.
\newblock Modeling Multi-turn Conversation with Deep Utterance Aggregation.
\newblock In \emph{Proceedings of the 27th International Conference on
  Computational Linguistics ({COLING} 2018)}.

\bibitem[{Zhang et~al.(2020{\natexlab{b}})Zhang, Wu, Zhou, Duan, Zhao, and
  Wang}]{zhang2020sg}
Zhang, Z.; Wu, Y.; Zhou, J.; Duan, S.; Zhao, H.; and Wang, R.
  2020{\natexlab{b}}.
\newblock SG-Net: Syntax-Guided Machine Reading Comprehension.
\newblock In \emph{AAAI}, 9636--9643.

\bibitem[{Zhu et~al.(2018{\natexlab{a}})Zhu, Wei, Qin, and Liu}]{haf}
Zhu, H.; Wei, F.; Qin, B.; and Liu, T. 2018{\natexlab{a}}.
\newblock Hierarchical Attention Flow for Multiple-Choice Reading
  Comprehension.
\newblock In \emph{Proceedings of the Thirty-Second {AAAI} Conference on
  Artificial Intelligence (AAAI 2018)}.

\bibitem[{Zhu et~al.(2018{\natexlab{b}})Zhu, Zhang, Li, Huang, and
  Zhao}]{lingke}
Zhu, P.; Zhang, Z.; Li, J.; Huang, Y.; and Zhao, H. 2018{\natexlab{b}}.
\newblock Lingke: A Fine-grained Multi-turn Chatbot for Customer Service.
\newblock In \emph{The 27th International Conference on Computational
  Linguistics: System Demonstrations (COLING DEMO 2018)}.

\end{thebibliography}
\bibliographystyle{aaai21}
\end{document}